\documentclass[journal]{IEEEtran}
\usepackage[mathlines,switch]{lineno}
\usepackage{graphicx}
\usepackage{amssymb}
\usepackage{amsmath}
\usepackage{enumerate}
\usepackage{algorithmic}
\usepackage{algorithm}
\usepackage[usenames]{color}
\usepackage{url}
\usepackage{multirow} 
\usepackage{booktabs}
\usepackage{threeparttable}
\usepackage{lineno}
\newtheorem{def1}{Definition}

\ifCLASSINFOpdf
\else
\fi
\begin{document}
%
\title{Memetic search for identifying critical nodes in sparse graphs}
%
%
%

\author{Yangming~Zhou, Jin-Kao~Hao*, Fred Glover
\thanks{Y. Zhou and J.K. Hao (Corresponding author) are with the Department of Computer Science, LERIA, Universit\'{e} d'Angers, 2 Boulevard Lavoisier, 49045 Angers, Cedex 01, France, J.K. Hao is also affiliated with the Institut Universitaire de France, 1 rue Descartes, 75231 Paris, Cedex 05, France (e-mail: yangming@info.univ-angers.fr; jin-kao.hao@univ-angers.fr).
F. Glover is with the University of Colorado, Leeds School of Business, Boulder, CO, USA (e-mail: glover@opttek.com).}}

\maketitle

\begin{abstract}
Critical node problems involve identifying a subset of critical nodes from an undirected graph whose removal results in optimizing a pre-defined measure over the residual graph. As useful models for a variety of practical applications, these problems are computational challenging. In this paper, we study the classic critical node problem (CNP) and introduce an effective memetic  algorithm for solving CNP. The proposed algorithm combines a double backbone-based crossover operator (to generate promising offspring solutions), a component-based neighborhood search procedure (to find high-quality local optima) and a rank-based pool updating strategy (to guarantee a healthy population). Specially, the component-based neighborhood search integrates two key techniques, i.e., two-phase node exchange strategy and node weighting scheme. The double backbone-based crossover extends the idea of general backbone-based crossovers. Extensive evaluations on 42 synthetic and real-world benchmark instances show that the proposed algorithm discovers 21 new upper bounds and matches 18 previous best-known upper bounds. We also demonstrate the relevance of our algorithm for effectively solving a variant of the classic CNP, called the cardinality-constrained critical node problem (CC-CNP). Finally, we investigate the usefulness of each key algorithmic component.
\end{abstract}


%
\IEEEpeerreviewmaketitle


\section{Introduction}
\label{Sec:Introduction}

Given an undirected graph $G = (V,E)$ with vertex (node) set $V$ and edge set $E$, \textsl{critical node problems} aim to delete a ``limited'' subset of nodes $S \subseteq V$ from $G$ such that a predefined connectivity metric over the residual graph $G[V \setminus S]$ (i.e., the sub-graph of $G$ induced by $V \setminus S$) is maximized or minimized. These deleted nodes in $S$ are commonly called \textsl{critical nodes}.

CNPs have natural applications in a number of fields, such as  network vulnerability assessment \cite{Shen2013, Chen2016a}, epidemic control \cite{Ventresca2014}, biological molecule studies \cite{Boginski2009, Tomaino2012}, network immunization \cite{Arulselvan2009b, Kuhlman2010}, network communications \cite{Commander2007}, network attacks \cite{Lozano2017} and social network analysis \cite{Borgatti2006, Leskovec2007, Fan2010}. For instance, in a social network, each node corresponds to a person, edges represent some type of interactions between the individuals (e.g., friendship or collaboration), and critical nodes correspond to the ``key players'' of the network (e.g., leaders of the organization or community) \cite{Borgatti2006}.

In this work, we are interested in the classic critical node problem (denoted as CNP hereinafter), which involves optimizing a pair-wise connectivity measure in the residual graph, i.e., minimizing the total number of connected node pairs. CNP is known to be NP-hard on general graphs \cite{Arulselvan2009b}, even if there are polynomially solvable special cases \cite{Shen2012}. The computational challenge and wide range of applications of CNP have motivated a variety of solution approaches in the literature, including exact algorithms and heuristic algorithms. Exact solution methods \cite{Arulselvan2009b, Summa2012, Veremyev2014a} guarantee the optimality of the solutions they find, but may fail on hard and large instances. Heuristic algorithms without guaranteed optimality of their solutions have also been studied to find good approximate solutions for large and hard instances within a reasonable computing time. For instance, an early heuristic starts with an independent set and uses a greedy criterion to remove vertices from the set \cite{Arulselvan2009b}. Another greedy algorithm using a modified depth-first search is proposed in \cite{Ventresca2015}. More recently, a number of metaheuristic algorithms have been reported for CNP, including iterated local search \cite{Aringhieri2016b,Zhou2017b}, variable neighborhood search \cite{Aringhieri2016b}, multi-start greedy algorithm \cite{Pullan2015}, greedy randomized adaptive search procedure with path relinking \cite{Purevsuren2016}, and genetic algorithm \cite{Aringhieri2016c}.

Neighborhood search plays a particularly important role in a metaheuristic search algorithm for CNP. In general, a neighborhood for CNP can be conveniently defined by the exchange (or swap) operation which exchanges a vertex in $S$ against a vertex in $V \setminus S$. However, this neighborhood has a quadratic size in terms of the number of nodes, making its exploration highly expensive.


To alleviate this difficulty and create an effective neighborhood search procedure, we propose a component-based neighborhood, which relies on a two-phase node exchange strategy and a node weighting technique. First, the two-phase node exchange strategy decomposes the exchange operation into two phases: a removal phase and an add phase, and performs them separately. Moreover, based on the fact that some swaps are irrelevant for optimizing the objective function, we constrain the exchange operations to some specific nodes (i.e., from large connected components in the residual graph). This constrained component-based neighborhood not only considerably reduces the number of candidate solutions to consider at each search iteration, but also makes the search more focused. Moreover, to make the node exchange operation efficient, we devise a node weighting technique to provide useful information for node selection within each exchange operation. Based on this component-based neighborhood, we introduce an effective local optimization procedure and apply it together with a double-backbone based crossover operator which can generate new promising offspring solutions from existing parent solutions. The whole algorithm (called MACNP for memetic algorithm for CNP), which also integrates a rank-based pool updating mechanism, proves to be highly effective for solving CNP. In addition, we extend the proposed algorithm to solve a cardinality constrained version of the classic CNP, i.e., the cardinality-constrained critical node problem (CC-CNP).  We summarize our main contributions as follows. 


\begin{itemize}
    \item First, the component-based neighborhood search procedure integrates two original ingredients, i.e., a two phase node exchange strategy and a node weighting scheme, which equips local optimization with a more focused and reduced neighborhood (Section \ref{SubSec:Component-based Neighborhood Search}). The double backbone-based crossover operator extends the idea of backbone-based crossovers by adopting a double backbone structure (Section \ref{SubSec:Double Backbone Based Crossover}). 
    \item The proposed MACNP algorithm yields highly competitive results compared with the state-of-the-art algorithms on both synthetic and real-world benchmarks. In particular, for the set of 16 synthetic instances, MACNP discovers 6 new upper bounds and matches the best known upper bounds for all 10 remaining instances. For the set of 26 real-world instances, MACNP attains 15 new upper bounds and matches 8 previous best-known upper bounds. Compared to the state-of-the-art algorithms, our algorithm also shows a superior performance on CC-CNP and achieves the best objective values on 39 out of 42 benchmark instances, yielding 22 new upper bounds.
\end{itemize}

The remainder of the paper is organized as follows. In Section \ref{Sec:Introduction}, we recapitulate the description of the critical node problem and indicate its relation to some other problems. In Section \ref{Sec:The Proposed Memetic Search for CNP}, we describe the proposed MACNP algorithm. In Section \ref{Sec:Computational Studies}, we present computational results for MACNP in comparison with the results of state-of-the-art algorithms. To show the generality of the proposed approach, we also verify its performance on the cardinality constrained critical node problem in Section \ref{Sec:Application to the Cardinality-Constrained Critical Node Problem}. In Section \ref{Sec:Discussion}, we experimentally analyze several key ingredients of the proposed approach to understand their impacts on the performance of the  algorithm. Concluding remarks are provided in the last section.

\section{Problem description and notation}
\label{Sec:Problem Description and Notations}

Critical node problems in a graph $G = (V,E)$ aim to delete a ``limited'' subset of nodes $S \subseteq V$ in order to maximize or minimize a pre-defined connectivity measure over the residual graph $G[V \setminus S]$. Once the critical nodes have been removed, the residual graph $G[V \setminus S]$ can be represented by a set of connected components $\mathcal{H} = \{\mathcal{C}_1,\mathcal{C}_2,\ldots,\mathcal{C}_T\}$, where a connected component $\mathcal{C}_i$ is a set of nodes such that all nodes in this set are mutually connected (reachable by some paths), and no two nodes in different sets are connected. 

Critical node problems have been extensively investigated in the last decade, and different connectivity measures have been studied according to the particular interests. These connectivity measures can be divided into three categories: 

\begin{itemize}
	\item [(i)] To optimize the pair-wise connectivity, i.e., the total number of pairs of nodes connected by a path in the residual graph \cite{Arulselvan2009b, Dinh2012, Addis2013, Veremyev2014b, Ventresca2015, Pullan2015, Addis2016, Purevsuren2016, Aringhieri2016b, Aringhieri2016c}.
	\item [(ii)] To optimize the size of the largest connected component in the residual graph \cite{Shen2012, Veremyev2014b, Pullan2015, Aringhieri2016c}.
	\item [(iii)] To optimize the total number of connected components in the residual graph \cite{Shen2012, Veremyev2014b, Aringhieri2016c}.
\end{itemize}

However, most studies in the literature have focused on the classic critical node problem (denoted as CNP), which aims to minimize the pair-wise connectivity measure \cite{Arulselvan2009b} and belongs to the first category mentioned above. Formally, given an integer $K$, CNP is to identify a subset $S \subseteq V$ where $|S| \leqslant K$, such that the following pair-wise connectivity objective $f(S)$ is minimized:

\begin{equation}\label{Equ:CNP Objective Function}
	f(S) = \sum_{i=1}^{T} \binom{|\mathcal{C}_i|}{2}
\end{equation}

where $T$ is the total number of connected components $\mathcal{C}_i$ in the residual graph $G[V\setminus S]$. 

CNP can be also considered as a problem of maximally fragmenting a graph and simultaneously minimizing the variance among the sizes of connected components in the residual graph. In other words, the resulting residual graph should be composed of a relatively large number of connected components while each connected component has a similar size \cite{Ventresca2015}. 

In this paper, we focus on solving this classic critical node problem. In Section \ref{Sec:Application to the Cardinality-Constrained Critical Node Problem}, we also show the applicability of our memetic algorithm to solve an important variant of CNP, i.e., the cardinality-constrained critical node problem (CC-CNP) \cite{Arulselvan2011}, which falls into the second category mentioned above.

CNP is closely related to a variety of other NP-hard optimization problems. For example, the $k$-cut problem \cite{Guttmann-Beck2000}, which is a popular graph partitioning problem. Given an undirected weighted graph, the $k$-cut problem is to find a minimum cost set of edges that separates the graph into $k$ connected components. Another similar problem is the minimum contamination problem, which  minimizes the expected size of contamination by removing a set of edges of at most a given cardinality \cite{Kumar2010}.

\section{The proposed memetic approach for CNP}
\label{Sec:The Proposed Memetic Search for CNP}

In this section, we present MACNP, an effective memetic algorithm for solving the classic critical node problem. The memetic framework is a powerful general method which has been successfully applied to solve many NP-hard problems, such as graph coloring \cite{Lu2010}, graph partition \cite{Benlic2011,Yuan2015}, maximum diversity \cite{Wu2013, Zhou2017} and quadratic knapsack \cite{Chen2016b}. The memetic framework combines population-based search and single-trajectory local search to achieve a suitable balance between search intensification and diversification.

\subsection{Solution representation and evaluation}
\label{SubSec:Solution Representation and Evaluation}

Given a graph $G = (V,E)$ and an integer $K$ (the maximum allowed number of nodes that can be removed), a feasible solution of CNP can be represented by $S = \{v_{S(1)},v_{S(2)},\ldots,v_{S(K)}\}$ ($1 \leqslant S(i) \neq S(j) \leqslant |V|$ for all $i \neq j$) where $S(l)$ ($1 \leqslant l \leqslant K$) is the index of a selected node in $S$. Therefore, the whole solution space $\Omega$ contains all possible subsets $S \subseteq V$ such that $|S| \leqslant K$.

According to Equation (\ref{Equ:CNP Objective Function}), for a feasible solution $S$, the corresponding objective function value $f(S)$ calculates the total number of node pairs still connected in the residual graph $G[V \setminus S]$. $f(S)$ can be computed with a modified depth-first search algorithm by identifying the connected components of a graph \cite{Hopcroft1973}, requiring $O(|V|+|E|)$ time. The modified depth-first search algorithm works as follows. It finds the connected components of a graph by performing the depth-first search on each connected component. Each new node visited is marked. When no more nodes can be reached along the edges from the marked nodes, a connected component is found. Then, an unvisited node is selected, and the process is repeated until the entire graph is explored.

\subsection{General scheme}
\label{SubSec:General Scheme}

The proposed MACNP algorithm is composed of four main procedures: a population initialization procedure, a component-based neighborhood search procedure, a double backbone-based crossover procedure and a rank-based pool updating procedure. MACNP starts from a set of distinct elite individuals which are obtained by the population initialization procedure (Section \ref{SubSec:Population Initialization}). At each generation, an offspring solution is generated by the double backbone-based crossover procedure (Section \ref{SubSec:Double Backbone Based Crossover}). This offspring solution is further improved by the component-based neighborhood search procedure (Section \ref{SubSec:Component-based Neighborhood Search}) and then considered for acceptance by the rank-based pool updating procedure (see Section \ref{SubSec:Rank-based Pool Updating}). The process is repeated until a stopping condition (e.g., time limit) is satisfied. The general framework of the proposed MACNP algorithm is presented in Algorithm \ref{Algorithm:MACNP} while its four procedures are respectively described in the following sections.

\begin{algorithm}
\begin{small}
 \caption{The proposed memetic algorithm for CNP}
 \label{Algorithm:MACNP}
 \begin{algorithmic}[1]
 	\STATE \sf \textbf{Input}: an undirected graph G = (V,E) and an integer $K$
 	\STATE \textbf{Output}: the best solution $S^*$ found so far\\
 	// build an initial population, Section \ref{SubSec:Population Initialization}\\
 	\STATE $P=\{S^1,S^2,\ldots,S^{p}\} \leftarrow$ PoolInitialize$()$
	\STATE $S^* = \arg \min\{f_1(S^i) : i = 1,2,\ldots,p\}$ \\
	\WHILE{a stopping condition is not reached}
		\STATE randomly select two parent solutions $S^i$ and $S^j$ from $P$\\
		// generate an offspring by crossover, Section \ref{SubSec:Double Backbone Based Crossover}\\
		\STATE $S' \leftarrow$ DoubleBackboneBasedCrossover$(S^i,S^j)$\\
		// perform a local search, Section \ref{SubSec:Component-based Neighborhood Search}\\
		\STATE $S' \leftarrow$ ComponentBasedNeighborhoodSearch$(S')$
		\IF{$f(S') < f(S^*)$} 
			\STATE $S^* = S'$
		\ENDIF
		\\
		// insert or discard the improved solution, Section \ref{SubSec:Rank-based Pool Updating}\\
		\STATE $P \leftarrow$ RankBasedPoolUpdating $(P,S')$ \\
	\ENDWHILE
 \end{algorithmic}
 \end{small}
 \end{algorithm}

\subsection{Population initialization}
\label{SubSec:Population Initialization}


Our MACNP algorithm starts its search with an initial population composed of diverse and high-quality solutions. To construct such a population, we first generate randomly a feasible solution (i.e., any set of at most $K$ nodes), and then we improve it by the component-based neighborhood search procedure described in Section \ref{SubSec:Component-based Neighborhood Search}. We insert the improved solution into the population if it is different from the existing individuals of the population. Otherwise, we modify the improved solution with the exchange operation until it becomes different from all existing individuals before inserting it into the population. We repeat the procedure $p$ times to fill the population with $p$ distinct solutions. 

\subsection{Component-based neighborhood search}
\label{SubSec:Component-based Neighborhood Search}

To ensure an effective local optimization, MACNP employs a fast and effective component-based neighborhood search (denoted by CBNS) procedure (see Algorithm \ref{Algorithm:Neighborhood Search}). CBNS integrates two key techniques, i.e., a two-phase node exchange strategy and a node weighting technique.

\begin{algorithm}
\begin{small}
 \caption{Component-based neighborhood search}
 \label{Algorithm:Neighborhood Search}
 \begin{algorithmic}[1]
 	\STATE \sf \textbf{Input}: a starting solution $S$
 	\STATE \textbf{Output}: the best solution $S^{*}$ found so far
 	\STATE $S^* \leftarrow S$\\
 	\STATE $iter \leftarrow 0$
	\WHILE{$iter < MaxIter$} 
		\STATE select a large component $c$ at random
		\STATE remove a node $u$ from component $c$ with the node weighting scheme\\
		\STATE $S \leftarrow S \cup \{u\}$\\
		\STATE $v \leftarrow \arg \min_{w \in S} \{f(S \setminus \{w\})-f(S)\}$\\
		\STATE $S \leftarrow S \setminus \{v\}$\\
		\IF{$f(S) < f(S^*)$} 
			\STATE $S^* \leftarrow S$\\
			\STATE $iter \leftarrow 0$\\
		\ELSE
			\STATE $iter \leftarrow iter + 1$\\
		\ENDIF
	\ENDWHILE
 \end{algorithmic}
\end{small}
\end{algorithm}

\subsubsection{Component-based neighborhood structure}
\label{SubSubSec:Component-based Neighborhood Structure}

The performance of a local search procedure greatly depends on its neighborhood structure for candidate solution exploration. A traditional neighborhood for CNP is defined by the conventional exchange operator which swaps a node $u \in S$ with a node $v \in V \setminus S$ \cite{Addis2016,Purevsuren2016,Ventresca2012}. For a given solution, this neighborhood yields $O(K(|V|-K))$ neighboring solutions. To evaluate a neighboring solution, no incremental technique is known and a full computation from scratch is required by running the modified depth-first search algorithm of complexity $O(|V|+|E|)$ \cite{Hopcroft1973}. Therefore, examining the whole neighborhood requires a time of $O(K(|V|-K)(|V|+|E|))$, which becomes too expensive when many local search iterations are performed (which is usually the case).

Very recently, two other refined neighborhoods have been proposed in \cite{Aringhieri2016b}. For a given $u \in S$, the first neighborhood aims to directly determine the node $v = \arg \max\{f(S)-f((S \setminus \{u\}) \cup \{v'\})\}$, $\forall v' \in V \setminus S$ so that the swap operation between $u$ and $v$ disconnects the graph as much as possible. For $v \in V \setminus S$, the second neighborhood tries to identify the node $u = \arg \min\{f((S \cup \{v\})\setminus \{u'\})-f(S)\}$. The computational complexity of examining these neighborhoods is $O(K(|V|+|E|))$ and $O((|V|-K)(|V|+|E|+ K \times degree(G)))$ respectively, where $degree(G)$ is the maximum node degree in $G$. Even if these neighborhoods are more computationally efficient compared to the traditional swap neighborhood, they are still expensive to explore within a local search algorithm. 

To overcome the limitation, we design an alternative and more efficient component-based neighborhood which is both smaller in size and more focused with respect to the optimization objective. Recall that CNP involves fragmenting the graph in order to minimize the total number of connected node pairs in the residual graph $G[V \setminus S]$. This can be achieved by fragmenting the largest connected components in the residual graph in order to obtain more homogeneous components, which helps to minimize the number of node pairs still connected. As a result, when exchanging a node $u \in S$ with a node $v \in V \setminus S$, it is preferable to consider $v \in V \setminus S$ from a large component (see Definition \ref{Def:Large Component} below) instead of a small component. Let ${L}$ be a predefined threshold to qualify large components. We consider only a subset of nodes $Z \subset V \setminus S$ such that $Z = \cup_{|\mathcal{C}_i| \geqslant L}\mathcal{C}_i$ as candidate nodes for exchanges.  Consequently, the neighborhood size is reduced to $K|Z|$, which is generally far smaller than $K(|V|-K)$ for reasonable ${L}$ values we used in this paper.

\begin{def1}[\textbf{large component}]\label{Def:Large Component}
A connected component in the residual graph $G[V\setminus S]$ qualifies as a large component if the number of its nodes is greater than the predefined threshold $L = (max\_n_c + min\_n_c)/2$, where $max\_n_c$ and $min\_n_c$ are respectively the number of nodes in the largest and smallest connected components in the residual graph $G[V\setminus S]$.
\end{def1}

\subsubsection{Two-phase node exchange strategy}
\label{SubSubSec:Two-Phase Node Exchange Strategy}

To further reduce the size of the above component-based neighborhood, we employ a two-phase node exchange strategy which relies on an extension of a candidate list strategy also called a neighborhood decomposition strategy or a successive filtration strategy \cite{Glover1993, Rangaswamy1998, Zhou2017}. The two-phase node exchange strategy breaks an exchange operation on a node pair into two distinct phases: a ``removal phase'' removes a node from the residual graph and an ``add phase''  adds a removed node back to the residual graph. This type of two-phase strategy is often used in conjunction with a candidate list approach in tabu search (see, e.g., \cite{GloverB}). In our case, the two-phase node exchange strategy first selects a component at random among the qualified large components and removes a node $v$ from the selected component with the node weighting scheme. For the node $u \in S$ to be moved to the residual graph, we select the node which minimally deteriorates the objective function. With the help of the two-phase node exchange strategy, the computational effort required to examine the candidate solutions greatly decreases. Consider the CNP instance `BA1000' with 1000 vertices and $K = 75$ as an example. Using the conventional exchange neighborhood requires consideration of $(1000-75) \times 75 = 69375$ candidate solutions. Instead, by adopting our two-phase node exchange strategy and only considering the qualified large connected components, only $75 \ll 69375$ candidate solutions need to be evaluated by our local search procedure because the process of removing a node from a connected component is performed regardless of its influence on the objective function. As we show in Section \ref{Sec:Computational Studies}, the component-based neighborhood with the two-phase node exchange strategy makes the search much more efficient.

The two-phase exchange strategy yields an efficient neighborhood search, in which the process of selecting a node $v$ to remove from $G[V \setminus S]$ is performed regardless of its influence on the objective function. The process can be finished in $O(T + nbr_v)$, where $T$ is the number of connected components in $G[V \setminus S]$ and $nbr_v$ is the length of the adjacency list of node $v$. Once a node is added into $S$, $S$ is an infeasible solution ($|S| = K+1$) and we need to remove a node $u$ from $S$, where we select $u$ to cause the minimum increase in the objective function. The evaluation of the increase in the objective function for each node in $S$ is performed by scanning the adjacency list of the node to determine if removing the node will re-connect some existing components to form a large component in the residual graph $G[V \setminus S]$. This operation requires time $O(K*nbr_u)$ where $nbr_u$ is the length of the adjacency list of node $u$.

\subsubsection{Node weighting scheme}
\label{SubSubSec:Node Weighting Scheme}

The node weighting technique is the second useful technique we adopted in the component-based neighborhood search. Weighting is a popular technique, which has been used in a number of heuristic algorithms, such as clause weighting for satisfiability problems \cite{Thornton2005}, edge weighting for the minimum vertex cover problem \cite{Cai2015}, and row weighting for the set cover problem \cite{Gao2015}. 

Our node weighting scheme works as follows. Each node of a large component is associated with a positive integer as its weight, initialized to 0. At each step, we randomly select a component $\mathcal{C}_i$ among the large connected components, and select the node $v$ in $\mathcal{C}_i$ with the largest weight (breaking ties in favor of the node with the largest degree) to move to $S$. Simultaneously, the weights of the remaining nodes in $\mathcal{C}_i$ are increased by one. Additionally, when a node $v \in \mathcal{C}_i$ is exchanged with a node $u \in S$, we set the weight of $u$ to 0.

With the help of the node weighting scheme, the ``hard to remove'' nodes will have larger weights, and thus have a higher chance to be considered for removal from the component in the following iterations. The node weighting technique helps the search to escape from potential local optima. Our node weighing scheme follows the general penalty idea for constraint satisfaction problems, which was first used in this setting in Morris's breakout method \cite{Morris1993}. We note that this scheme is also an instance of a tabu search frequency-based memory (see, e.g., the six frequency-based memory classes proposed earlier in \cite{GloverB} and their refinements in \cite{GloverE}). To the best of our knowledge, it is the first time that a node weight learning technique is applied to a heuristic procedure for CNP. 

\subsection{Double backbone-based crossover}
\label{SubSec:Double Backbone Based Crossover}

Crossover is another important ingredient of the MACNP algorithm. It should be noted that the meaning of “crossover” has changed from the genetic conception adopted in the early formulation of memetic algorithms. The modern conception embraces the principle of structured combinations introduced in \cite{GloverC}, where solutions are combined by domain specific heuristics that map them into new solutions faithful to the structure of the problems considered. (A similar evolution in the notion of crossover has been occurring within genetic algorithms to incorporate the notion of structured combinations, although often incompletely.) As observed in \cite{Hao2012}, a successful crossover should be able to generate promising offspring solutions by inheriting good properties of the parents and introducing useful new features, while respecting the domain specific structure of the problem context . The concept of backbone has been used to design some successful crossover operators for subset selection problems \cite{Wu2013, Zhou2017}. The critical node problem being a typical subset selection problem, we adopt the backbone idea and design a double backbone-based crossover operator to create structured combinations as follows.

Let $S^1$ and $S^2$ be two solutions of CNP. According to $S^u$ and $S^v$, we divide the set of elements $V$ into three subsets of common elements, unique elements and unrelated elements, as shown in Definition \ref{Def:Common Elements}, \ref{Def:Exclusive Elements} and \ref{Def:Excluding Elements} respectively.

\begin{def1}[\textbf{common elements}]\label{Def:Common Elements}
The set of common elements $X_A$ is the set of elements of $V$ shared by $S^1$ and $S^2$, i.e., $X_A = S^1 \cap S^2$.
\end{def1}

\begin{def1}[\textbf{exclusive elements}]\label{Def:Exclusive Elements}
The set of exclusive elements $X_B$ is the set of elements of $V$ shared by either $S^1$ or $S^2$, i.e., $X_B = (S^1 \cup S^2) \setminus (S^1 \cap S^2)$ (the symmetric difference of $S^1$ and $S^2$).
\end{def1}

\begin{def1}[\textbf{excluding elements}]\label{Def:Excluding Elements}
The set of excluding elements $X_C$ is the set of elements of $V$ which are not included in $S^1$ and $S^2$, i.e., $X_C = V \setminus (S^1 \cup S^2)$.
\end{def1}

From two parent solutions $S^1$ and $S^2$ randomly selected from the population $P$, an offspring solution $S^0$ is constructed in three phases: (i) create a partial solution by inheriting all common elements (i.e., the first backbone), i.e., $S^0 \leftarrow X_A$; (ii) add exclusive elements (i.e., the second backbone) into the partial solution in a probabilistic way. That is, for each exclusive element, we add it into $S^0$ with probability $p_0$ ($0 < p_0 < 1$), otherwise we discard it; (iii) repair the partial solution  structurally until a feasible solution is achieved. Specifically, if $|S^0| < K$, we randomly add some elements to $S^0$ from a random  large connected component in the residual graph; Otherwise we greedily remove some elements from $S^0$ until $|S^0| = K$. The elements added in the first two phases form the whole backbone of the parent solutions. Therefore, the double backbones are composed of $|X_A|$ common elements (i.e., the first backbone) and about $p_0*|X_B|$ exclusive elements (i.e., the second backbone).

This double backbone-based crossover operator shares similar ideas with the crossovers proposed in \cite{Wu2013, Zhou2017}, i.e., directly inheriting all common elements from its parent solutions (see Definition \ref{Def:Common Elements}). However, our double backbone based crossover operator distinguishes itself from these crossovers by adopting the double backbone structure. That is, it also directly inherits some exclusive elements (see Definition \ref{Def:Exclusive Elements}) with a selection probability $p_0$. This strategy of combining solutions by introducing elements beyond their union is shared with the approach of exterior path relinking \cite{GloverD}, which likewise has recently been found effective in discrete optimization.

\subsection{Rank-based pool updating}
\label{SubSec:Rank-based Pool Updating}

Each offspring solution is submitted for improvement by the component-based neighborhood search procedure presented in Section \ref{SubSec:Component-based Neighborhood Search}. Then we use a rank-based pool updating strategy to decide whether the improved offspring solution $S^0$ should be accepted in the population. This pool updating strategy resorts to a score function to evaluate each individual. The score function not only considers the quality of the offspring but also its average distance to other individuals in the population. This strategy is inspired by the population management strategies presented in \cite{Lu2010, Chen2016b, Zhou2017}.

The rank-based pool updating strategy applied in our algorithm is described in Algorithm \ref{Algorithm:Rank-based Pool Updating Strategy}. At first, we temporarily insert $S^0$ to the population $P$ (line 3 of Alg.\ref{Algorithm:Rank-based Pool Updating Strategy}), then we evaluate all individuals of the population according to the score function \cite{Zhou2017} (lines 4-8 of Alg.\ref{Algorithm:Rank-based Pool Updating Strategy}) and identify the worst solution $S^w$ with the largest $Score$ value (line 9 of Alg.\ref{Algorithm:Rank-based Pool Updating Strategy}). Finally, if $S^0$ is different from $S^w$, we replace $S^w$ by $S^0$. Otherwise, we discard $S^0$ (lines 10-12 of Alg.\ref{Algorithm:Rank-based Pool Updating Strategy}).

\begin{algorithm}
\begin{small}
 \caption{Rank-based pool updating strategy}
 \label{Algorithm:Rank-based Pool Updating Strategy}
 \begin{algorithmic}[1]
     \STATE \sf \textbf{Input}: a population $P$ and an improved solution $S^0$
     \STATE \textbf{Output}: a new population $P$
     \STATE $P' \leftarrow P \cup \{S^0\}$\\
     \STATE $i \leftarrow 0$
     \WHILE{$i \leqslant p$} 
         \STATE Evaluate individual $S^i$ according to the score function
         \STATE $i \leftarrow i + 1$
     \ENDWHILE
     \STATE Identify the worst solution $S^w$ in population $P'$\\
      i.e., $w \leftarrow \max_{j \in \{0,1,\ldots,p\}}Score(S^j,P')$
     \IF{$w \neq 0$} 
         \STATE Replace $S^w$ with $S^0$, i.e., $P \leftarrow P' \setminus \{S^w\}$
     \ENDIF
 \end{algorithmic}
\end{small}
\end{algorithm}

\subsection{Computational complexity of MACNP}
\label{SubSec:Computational Complexity of the Proposed MACNP}

To analyze the computational complexity of the proposed MACNP algorithm, we consider the main steps in one generation in the main loop of Algorithm \ref{Algorithm:MACNP}.

As displayed in Algorithm \ref{Algorithm:MACNP}, at each generation, our MACNP algorithm consists of four subroutines: parent selection, double backbone-based crossover, component-based neighborhood search and rank-based pool updating. The parent selection is very simple and takes time $O(1)$. The double backbone-based crossover operator can be realized in time $O(|V|K^2)$. The computational complexity of the component-based neighborhood search is $O(K(|V|+|E|)MaxIter$, where $MaxIter$ is the maximum allowable number of iterations without improvement. The rank-based pool updating can be achieved in time $O(p(K^2+p))$, where $p$ is the population size. Hence, for each generation, the total complexity of MACNP  is $O(|V|K^2+K(|V|+|E|)MaxIter)$.

\section{Computational studies}
\label{Sec:Computational Studies}
This section presents computational studies to evaluate the performance of our MACNP algorithm and compare it with state-of-the-art algorithms.

\subsection{Benchmark instances}
\label{SubSec:Benchmark Instances}

Our computational studies were based on two benchmarks.\footnote{Both synthetic and real-world benchmark instances are publicly available at \url{http://www.di.unito.it/~aringhie/cnp.html}.}

\textbf{Synthetic benchmark} was originally presented in \cite{Ventresca2012} and contains 16 instances classified into four categories: Barabasi-Albert (BA) graphs, Erdos-Renyi (ER) graphs, Forest-Fire (FF) graphs and Watts-Strogatz (WS) graphs.

\textbf{Real-world benchmark} was first presented in \cite{Aringhieri2016c} and consists of 26 real-world graphs from various practical applications, including protein interaction, the electronic circuit, flight network, train network, electricity distribution network, social network and etc.

It is worth noting that both the benchmark instances are all sparse graphs. We use an indicator $\beta = 2|E|/(|V|(|V|+1))$ ($0 < \beta \leqslant 1$) to measure the sparse degree of an instance, and we observe that $\beta \leqslant 0.045$ holds for all instances. 

\subsection{Experimental settings}
\label{SubSec:Experimental Setttings}

The proposed algorithm\footnote{The code of our MACNP algorithm will be made available at \url{http://www.info.univ-angers.fr/~hao/cnps.html}} is implemented in the C++ programming language and complied with gcc 4.1.2 and flag `-O3'. All the experiments were carried out on a computer equipped with an Intel E5-2670 processor with 2.5 GHz and 2 GB RAM operating under the Linux system. Without using any compiler flag, running the well-known DIMACS machine benchmark procedure dfmax.c\footnote{dfmax: \url{ftp://dimacs.rutgers.edu/pub/dsj/clique}} on our machine requires 0.19, 1.17 and 4.54 seconds to solve the benchmark graphs r300.5, r400.5 and r500.5 respectively. Our computational results were obtained by running the MACNP algorithm with the parameter settings provided in Table \ref{Tab:Parameter Settings}. Given that the benchmark instances have different structures, it is difficult to obtain a set of parameter values which yield uniformly the best result on all instances. To determine these values, we evaluate the performance of the algorithm for each parameter by varying the chosen parameter within a reasonable range, while fixing the other parameters as the default value of Table \ref{Tab:Parameter Settings}.

\begin{table}[!ht]
\caption{The parameter settings of the proposed MACNP algorithm.}
\label{Tab:Parameter Settings}
\vskip -0.2in
\begin{center}
\begin{scriptsize}
\begin{tabular}{lllc}
\toprule[0.75pt]
Parameter & Description & Value & Section\\
\midrule[0.5pt]
$t_{max}$ & time limit (seconds) & 3600 & \ref{SubSec:General Scheme}\\
$p$ & population size    & 20 & \ref{SubSec:Population Initialization}\\
$MaxIter$ & max no improvement iteration in CBNS & 1000 & \ref{SubSec:Component-based Neighborhood Search}\\
$p_0$ & selection probability & 0.85 & \ref{SubSec:Double Backbone Based Crossover}\\
$\beta$ & weighting coefficient & 0.6 & \ref{SubSec:Rank-based Pool Updating}\\
\bottomrule[0.75pt]
\end{tabular}
\end{scriptsize}
\end{center}
\vskip -0.2in
\end{table}

For our experiments, we adopted a cutoff time as the stopping condition, which is a standard practice for solving CNPs \cite{Aringhieri2016b,Aringhieri2016c,Purevsuren2016,Zhou2017b}. Given the stochastic nature of the proposed algorithm, the algorithm was independently executed 30 times on each test instance like \cite{Ventresca2012}. 

To analyze the experimental results, we resort to the well-known two-tailed sign test \cite{Demvsar2006} to check the significant difference on each comparison indicator between the compared algorithms. When two algorithms are compared, the corresponding null-hypothesis is that the algorithms are equivalent. The null-hypothesis is accepted if and only if each algorithm wins on approximately $X/2$ out of $X$ instances. Since tied matches support the null-hypothesis, we split them evenly between the two compared algorithms, i.e., each one receives the value 0.5. At a significance level of 0.05, the Critical Values (CV) of the two-tailed sign test are respectively $CV_{0.05}^{16} = 12$ and $CV_{0.05}^{20} = 18$ when the number of instances in each benchmark is $X = 16$ and $X = 26$. Consequently, Algorithm A is significantly better than algorithm B if A wins at least $CV_{0.05}^X$ instances for a benchmark of $X$ instances.

\subsection{Performance of the MACNP algorithm}
\label{SubSec:Performance of the MACNP Algorithm}

\begin{table}[!htbp]
\caption{Performance of MACNP on synthetic and real-world benchmarks.}
\label{Tab:Computational Results on Synthetic and Realworld Benchmarks With 3600 Seconds}
\vskip -0.2in
\begin{center}
\begin{scriptsize}
\begin{threeparttable}
\begin{tabular}{lrrrrrc}
\toprule[0.75pt]
Instance 	 & $K$ & $KBV$ & $\Delta f_{best}$ & $\Delta f_{avg}$ & $t_{avg}$ & \#steps\\
\midrule[0.5pt]
BA500&50&195\tnote{$\ast$}&0&0.0&0.0&$5.8 \times 10^2$\\
BA1000&75&558\tnote{$\ast$}&0&0.0&0.3&$6.4 \times 10^3$\\
BA2500&100&3704\tnote{$\ast$}&0&0.0&0.7&$8.6 \times 10^3$\\
BA5000&150&10196\tnote{$\ast$}&0&0.0&6.5&$3.5 \times 10^4$\\
ER235&50&295\tnote{$\ast$}&0&0.0&7.1&$2.0 \times 10^5$\\
ER466&80&1524&0&0.0&28.5&$9.5 \times 10^5$\\
ER941&140&5012&0&2.1&458.5&$1.2 \times 10^7$\\
ER2344&200&959500&-57002&-37160.5&2284.8&$1.5 \times 10^7$\\
FF250&50&194\tnote{$\ast$}&0&0.0&0.0&$2.3 \times 10^3$\\
FF500&110&257\tnote{$\ast$}&0&0.0&0.4&$8.7\times 10^3$\\
FF1000&150&1260\tnote{$\ast$}&0&0.0&84.9&$2.6 \times 10^5$\\
FF2000&200&4545\tnote{$\ast$}&0&0.7&107.6&$3.3 \times 10^5$\\
WS250&70&3101&-18&-11.6&1140.5&$6.1 \times 10^7$\\
WS500&125&2078&-6&4.6&179.3&$2.4 \times 10^6$\\
WS1000&200&113638&-3831&10044.6&2675.0&$1.8 \times 10^7$\\
WS1500&265&13167&-69&88.1&1012.2&$7.3 \times 10^6$\\
\midrule[0.5pt]
Bovine&3&268&0&0.0&0.0&$6.8 \times 10^1$\\
Circuit&25&2099&0&0.0&0.2&$1.7 \times 10^4$\\
Ecoli&15&806&0&0.0&0.0&$6.2 \times 10^2$\\
USAir97&33&4336&0&0.0&756.5&$2.5 \times 10^6$\\
humanDi&52&1115&0&0.0&0.6&$1.7 \times 10^4$\\
TreniR&26&920&-2&-2.0&0.3&$2.2 \times 10^4$\\
EU\_fli &119&349927&-1659&1730.0&232.6&$1.1 \times 10^5$\\
openfli&186&28671&-1829&33.3&2093.7&$2.9 \times 10^6$\\
yeast1&202&1414&-2&-2.0&21.7&$1.0 \times 10^5$\\
H1000&100&328817&-22468&-18190.5&2137.5&$2.5 \times 10^7$\\
H2000&200&1309063&-65204&-45567.4&2861.9&$1.2 \times 10^7$\\
H3000a&300&3005183&-160790&-120401.3&3280.7&$1.1 \times 10^7$\\
H3000b&300&2993393&-152123&-108306.0&3252.9&$1.2 \times 10^7$\\
H3000c&300&2975213&-136784&-105864.5&3307.5&$1.1 \times 10^7$\\
H3000d&300&2988605&-157294&-96042.3&3250.9&$1.1 \times 10^7$\\
H3000e&300&3001078&-153169&-113552.3&3437.4&$1.2 \times 10^7$\\
H4000&400&5403572&-250595&-136196.5&2907.0&$6.6 \times 10^6$\\
H5000&500&8411789&-439264&-316976.4.7&3226.6&$6.3 \times 10^6$\\
powergr&494&16099&-237&-197.5&1286.4&$1.8 \times 10^6$\\
OClinks&190&614504&-2201&40.0&584.6&$4.5 \times 10^5$\\
faceboo&404&420334&222828&319102.6&2978.5&$2.2 \times 10^6$\\
grqc&524&13736&-140&-106.8&871.8&$9.2 \times 10^5$\\
hepth&988&114382&-7985&-4726.4&3442.0&$3.1 \times 10^6$\\
hepph&1201&7336826&1291861&2033389.3&3376.3&$1.4 \times 10^6$\\
astroph&1877&54517114&7551852&8030784.1&1911.4&$3.9 \times 10^5$\\
condmat&2313&2298596&7155765&7763211.8&1779.5&$4.9 \times 10^5$\\
\bottomrule[0.75pt]
\end{tabular}
\begin{tablenotes}
\tiny
     \item[$\ast$] Optimal results obtained by exact algorithm \cite{Summa2012} within 5 days.
\end{tablenotes}
\end{threeparttable}
\end{scriptsize}
\end{center}
\vskip -0.2in
\end{table}

Table \ref{Tab:Computational Results on Synthetic and Realworld Benchmarks With 3600 Seconds} shows the computational results for MACNP on the synthetic and real-world benchmarks under the time limit $t_{max} = 3600$ seconds. Columns 1-3 respectively describe for each instance its name (Instance), the number of critical nodes ($K$) and the known best objective value ($KBV$) reported in the literature. Columns 4-8 report the detailed results of MACNP, including the difference between the best objective value $f_{best}$ and its known best value $KBV$ (i.e., $\Delta f_{best} = f_{best}-KBV$), the difference between the average objective value $f_{avg}$ and $KBV$ (i.e., $\Delta f_{avg} = f_{avg}-KBV$), the average time to attain the objective value ($t_{avg}$) and the average number of steps (i.e., exchanges) to achieve the objective value (\#step).

From Table \ref{Tab:Computational Results on Synthetic and Realworld Benchmarks With 3600 Seconds}, we observe that MACNP is able to attain the best objective values for all 16 benchmark instances while yielding in particular 5 new upper bounds (see $\Delta f_{best} < 0$ in Table \ref{Tab:Computational Results on Synthetic and Realworld Benchmarks With 3600 Seconds}). For instances ER2344 and WS250, our average objective values are also better than the previously best known upper bound (see $\Delta f_{avg} < 0$ in Table \ref{Tab:Comparative Performance on Weighting Scheme}). To the best of our knowledge, our MACNP algorithm is the first heuristic which reaches the optimal solution 4545 of FF2000. The average time to find the optimal value 4545 is 107.6, which is far less than 5 days by the exact algorithm \cite{Summa2012} (as reported in \cite{Aringhieri2016b}). For the real-world benchmark, MACNP also shows a highly competitive performance and achieves the best objective value on 22 out of 26 instances, yielding 17 new upper bounds (see $\Delta f_{best} < 0$ in Table \ref{Tab:Computational Results on Synthetic and Realworld Benchmarks With 3600 Seconds}) and matches 5 previous upper bounds (see $\Delta f_{best} = 0$ in Table \ref{Tab:Computational Results on Synthetic and Realworld Benchmarks With 3600 Seconds}). Also, the average objective value achieved by our MACNP algorithm is better than the previous upper bound for 14 instances (see $\Delta f_{avg} < 0$ in Table \ref{Tab:Computational Results on Synthetic and Realworld Benchmarks With 3600 Seconds}). However, MACNP failed to attain the known best value for three large instances (hepph, astroph,and condmat) within the time limit $t_{max} = 3600$ seconds. Indeed, this time limit is too short for the population-based MACNP algorithm to converge. Note that in \cite{Aringhieri2016c}, a large time limit of  $t_{max} = 16000$ seconds was used. When we re-ran our MACNP algorithm under this condition, MACNP managed to find better solutions, including two new upper bounds (see results displayed in italic format in Table \ref{Tab:Comparisons of MACNP With the State-of-the-art Algorithms on Synthetic and Realworld Benchmarks}). This experiment  demonstrates the effectiveness of our MACNP algorithm for solving the CNP on both the synthetic and real-world benchmarks.

\subsection{Comparison with the state-of-the-art algorithms}
\label{SubSec:Comparison With Other State-of-the-art Algorithms}

To further assess the performance of our MACNP algorithm, we carried out detailed comparisons between MACNP and  state-of-the-art heuristic algorithms. We consider 7 reference algorithms, including the dynamic restarting greedy algorithms (Greedy3d and Greedy4d) \cite{Addis2016}, iterated local search (ILS) \cite{Aringhieri2016b}, variable neighborhood search (VNS) \cite{Aringhieri2016b}, genetic algorithm (GA) \cite{Aringhieri2016c}, multi-start greedy algorithm (CNA1) \cite{Pullan2015} and a fast heuristic (FastCNP) \cite{Zhou2017b}.

Since the source codes of CNA1 and FastCNP are available to us, we first make a detailed comparison between  MACNP and these two reference algorithms. To make a fair comparison, all the three algorithms were run on our platform with the same time limit $t_{max} = 3600$ seconds, and each algorithm was executed 30 trials to solve each instance. The comparative results are shown in Table \ref{Tab:Comparative Performance of MACNP with Two Best-performing Algorithms CNA1 and FastCNP}. In this table, the first column provides the name of each instance (Instance), Columns 2-5 report the results of the CNA1 algorithm, including the best objective value ($f_{best}$), the average objective value ($f_{avg}$), the average time to attain the objective value ($t_{avg}$) and the number of steps to achieve the objective value (\#step). Correspondingly, columns 6-9 and columns 10-13 respectively represent the results of algorithms FastCNP and MACNP. The best values of the compared results are in bold, and when the same best objective values are achieved, fewer steps are underlined (which indicates a better performance in terms of computational efficiency). In addition, we  give the number of instances (wins) for which our algorithm obtained a better performance (i.e., $f_{best}$ and $f_{avg}$) compared to the corresponding algorithms. The win values for indicators $t_{avg}$ and \#steps are meaningless and are marked by `*'.

\begin{table*}[!htbp]
\caption{Comparative performance of the proposed MACNP with CNA1 and FastCNP on synthetic and real-world benchmarks.}
\label{Tab:Comparative Performance of MACNP with Two Best-performing Algorithms CNA1 and FastCNP}
\vskip -0.4in
\begin{center}
\begin{scriptsize}
\begin{threeparttable}
\begin{tabular}{lrrrccrrrccrrrc}
\toprule[0.75pt]
\multicolumn{1}{c}{} & \multicolumn{4}{c}{CNA1} && \multicolumn{4}{c}{FastCNP} && \multicolumn{4}{c}{MACNP}\\
\cmidrule[0.5pt]{2-5} \cmidrule[0.5pt]{7-10} \cmidrule[0.5pt]{12-15}
Instance&$f_{best}$&$f_{avg}$&$t_{avg}$&\#steps &&$f_{best}$&$f_{avg}$&$t_{avg}$&\#steps &&$f_{best}$&$f_{avg}$&$t_{avg}$&\#steps\\
\midrule[0.5pt]
BA500&\textbf{195}&\textbf{195.0}&2.5&$1.4 \times 10^5$&&\textbf{195}&\textbf{195.0}&$<0.1$&$4.1 \times 10^3$&&\textbf{195}&\textbf{195.0}&$<0.1$&\underline{$5.8 \times 10^2$}\\
BA1000&\textbf{558}&558.7&5.4&$1.1 \times 10^5$&&\textbf{558}&\textbf{558.0}&29.8&$1.6 \times 10^6$&&\textbf{558}&\textbf{558.0}&0.3&\underline{$6.4 \times 10^3$}\\
BA2500&\textbf{3704}&\textbf{3704.0}&1.7&$1.4 \times 10^4$&&\textbf{3704}&3710.6&649.9&$1.7 \times 10^7$&&\textbf{3704}&\textbf{3704.0}&0.7&\underline{$8.6 \times 10^3$}\\
BA5000&\textbf{10196}&\textbf{10196.0}&103.7&$3.8 \times 10^5$&&\textbf{10196}&10201.4&104.9&$1.5 \times 10^6$&&\textbf{10196}&\textbf{10196.0}&6.5&\underline{$3.5 \times 10^4$}\\
ER235&\textbf{295}&\textbf{295.0}&6.8&$1.0 \times 10^6$&&\textbf{295}&\textbf{295.0}&11.7&$4.2 \times 10^6$&&\textbf{295}&\textbf{295.0}&7.1&\underline{$2.1 \times 10^5$}\\
ER466&\textbf{1524}&\textbf{1524.0}&825.4&$7.2 \times 10^7$&&\textbf{1524}&\textbf{1524.0}&364.9&$7.0 \times 10^7$&&\textbf{1524}&\textbf{1524.0}&28.5&\underline{$9.5 \times 10^5$}\\
ER941&5114&5177.4&1606.1&$6.7 \times 10^7$&&\textbf{5012}&\textbf{5013.3}&1516.7&$1.4 \times 10^8$&&\textbf{5012}&5014.1&458.5&$1.2 \times 10^7$\\
ER2344&996411&1008876.4&1379.2&$8.8 \times 10^6$&&953437&979729.2&1793.8&$3.4 \times 10^7$&&\textbf{902498}&\textbf{922339.5}&2284.8&$1.5 \times 10^7$\\
FF250&\textbf{194}&\textbf{194.0}&158.0&$2.2 \times 10^7$&&\textbf{194}&\textbf{194.0}&1.9&$4.6 \times 10^5$&&\textbf{194}&\textbf{194.0}&$<0.1$&\underline{$2.3 \times 10^3$}\\
FF500&263&265.0&197.5&$9.4 \times 10^6$&&\textbf{257}&258.4&55.1&$5.3 \times 10^6$&&\textbf{257}&\textbf{257.0}&0.4&$8.7 \times 10^3$\\
FF1000&1262&1264.2&1743.0&$3.8 \times 10^7$&&\textbf{1260}&1260.8&23.6&$1.0 \times 10^6$&&\textbf{1260}&\textbf{1260.0}&84.9&$2.6 \times 10^5$\\
FF2000&4548&4549.4&1571.0&$2.4 \times 10^7$&&4546&4558.3&1160.8&$2.8 \times 10^7$&&\textbf{4545}&\textbf{4545.7}&107.6&$3.3 \times 10^5$\\
WS250&3415&3702.8&1424.4&$7.6 \times 10^7$&&3085&3196.4&1983.5&$4.2 \times 10^8$&&\textbf{3083}&\textbf{3089.4}&1140.5&$6.1 \times 10^7$\\
WS500&2085&2098.7&1581.4&$1.5 \times 10^8$&&\textbf{2072}&2083.3&1452.6&$2.7 \times 10^8$&&\textbf{2072}&\textbf{2082.6}&179.3&$2.4 \times 10^6$\\
WS1000&141759&161488.0&116.5&$1.2 \times 10^6$&&123602&127493.4&2120.2&$6.3 \times 10^7$&&\textbf{109807}&\textbf{123682.6}&2675.0&$1.8 \times 10^7$\\
WS1500&13498&13902.5&1787.2&$5.7 \times 10^7$&&13158&13255.7&1554.9&$8.8 \times 10^7$&&\textbf{13098}&\textbf{13255.1}&1012.2&$7.3 \times 10^6$\\
\midrule[0.5pt]
wins &12.5 & 13.0 & * & * & & 10.5 & 13.0 & * & * & & * & * &* &*\\
\midrule[0.5pt]
Bovine&\textbf{268}&\textbf{268.0}&$<0.1$&$3.0 \times 10^2$&&\textbf{268}&\textbf{268.0}&$<0.1$&$2.4 \times 10^3$&&\textbf{268}&\textbf{268.0}&$<0.1$&\underline{$6.8 \times 10^1$}\\
Circuit&\textbf{2099}&\textbf{2099.0}&0.3&$6.8 \times 10^4$&&\textbf{2099}&\textbf{2099.0}&1.2&$5.9 \times 10^5$&&\textbf{2099}&\textbf{2099.0}&0.2&\underline{$1.7 \times 10^4$}\\
E.coli&\textbf{806}&\textbf{806.0}&$<0.1$&$1.3 \times 10^3$&&\textbf{806}&\textbf{806.0}&$<0.1$&$7.8 \times 10^3$&&\textbf{806}&\textbf{806.0}&$<0.1$&\underline{$6.3 \times 10^2$}\\
USAir97&\textbf{4336}&\textbf{4336.0}&254.9&$1.1 \times 10^7$&&\textbf{4336}&\textbf{4336.0}&90.8&$8.6 \times 10^6$&&\textbf{4336}&\textbf{4336.0}&756.5&\underline{$2.5 \times 10^6$}\\
HumanDi&\textbf{1115}&\textbf{1115.0}&5.8&$4.8 \times 10^5$&&\textbf{1115}&\textbf{1115.0}&2.5&$4.3 \times 10^5$&&\textbf{1115}&\textbf{1115.0}&0.6&\underline{$1.7 \times 10^4$}\\
TreniR&\textbf{918}&\textbf{918.0}&1.3&$4.3 \times 10^5$&&\textbf{918}&\textbf{918.0}&2.4&$1.4 \times 10^6$&&\textbf{918}&\textbf{918.0}&0.3&\underline{$2.2 \times 10^4$}\\
EU\_fli &\textbf{348268}&\textbf{348347.0}&914.8&$1.6 \times 10^6$&&\textbf{348268}&348697.7&1495.0&$6.0 \times 10^6$&&\textbf{348268}&351657.0&232.6&\underline{$1.1 \times 10^5$}\\
openfli&29300&29815.3&1835.0&$7.4 \times 10^6$&&28834&29014.4&499.3&$3.5 \times 10^6$&&\textbf{26842}&\textbf{28704.3}&2093.7&$2.9 \times 10^6$\\
yeast&1413&1416.3&1461.9&$1.1 \times 10^7$&&\textbf{1412}&\textbf{1412.0}&252.0&$2.8 \times 10^6$&&\textbf{1412}&\textbf{1412.0}&21.7&$1.0 \times 10^5$\\
H1000&314152&317805.7&1412.4&$2.5 \times 10^7$&&314964&316814.8&1821.6&$6.7 \times 10^7$&&\textbf{306349}&\textbf{310626.5}&2137.5&$2.5 \times 10^7$\\
H2000&1275968&1292400.4&1200.0&$8.6 \times 10^6$&&1275204&1285629.1&1620.1&$2.5 \times 10^7$&&\textbf{1243859}&\textbf{1263495.6}&2861.9&$1.2 \times 10^7$\\
H3000a&2911369&2927312.0&1598.5&$7.4 \times 10^6$&&2885588&2906965.5&2041.5&$1.8 \times 10^7$&&\textbf{2844393}&\textbf{2884781.7}&3280.7&$1.1 \times 10^7$\\
H3000b&2907643&2927330.5&963.3&$4.3 \times 10^6$&&2876585&2902893.9&1596.2&$1.3 \times 10^7$&&\textbf{2841270}&\textbf{2885087.0}&3252.9&$1.2 \times 10^7$\\
H3000c&2885836&2917685.8&1142.6&$4.7 \times 10^6$&&2876026&2898879.3&1927.9&$1.7 \times 10^7$&&\textbf{2838429}&\textbf{2869348.5}&3307.5&$1.1 \times 10^7$\\
H3000d&2906121&2929569.2&1463.9&$5.7 \times 10^6$&&2894492&2907485.4&2005.4&$1.7 \times 10^7$&&\textbf{2831311}&\textbf{2892562.7}&3250.9&$1.1 \times 10^7$\\
H3000e&2903845&2931806.8&1489.4&$6.1 \times 10^6$&&2890861&2911409.3&1993.0&$1.6 \times 10^7$&&\textbf{2847909}&\textbf{2887525.7}&3437.4&$1.2 \times 10^7$\\
H4000&5194592&5233954.5&1749.1&$5.3 \times 10^6$&&5167043&5190883.7&1954.2&$1.2 \times 10^7$&&\textbf{5044357}&\textbf{5137528.3}&2907.0&$6.6 \times 10^6$\\
H5000&8142430&8212165.9&1342.5&$3.0 \times 10^6$&&8080473&8132896.2&2009.3&$8.8 \times 10^6$&&\textbf{7972525}&\textbf{8094812.6}&3226.6&$6.3 \times 10^6$\\
powergr&16158&16222.1&1532.2&$5.6 \times 10^6$&&15982&16033.5&1610.3&$9.8 \times 10^6$&&\textbf{15862}&\textbf{15901.5}&1286.4&$1.8 \times 10^6$\\
Oclinks&\textbf{611326}&614858.5&990.9&$2.5 \times 10^6$&&611344&616783.0&713.1&$3.3 \times 10^6$&&612303&\textbf{614544.0}&584.6&$4.5 \times 10^5$\\
faceboo&701073&742688.0&2234.4&$4.8 \times 10^6$&&692799&765609.8&3132.9&$1.5 \times 10^7$&&\textbf{643162}&\textbf{739436.6}&2978.5&$2.2 \times 10^6$\\
grqc&15522&15715.7&2201.1&$5.2 \times 10^6$&&13616&13634.8&2002.0&$7.8 \times 10^6$&&\textbf{13596}&\textbf{13629.2}&871.8&$9.2 \times 10^5$\\
hepth&130256&188753.7&2135.6&$2.1 \times 10^6$&&108217&109889.5&2765.3&$5.3 \times 10^6$&&\textbf{106397}&\textbf{109655.6}&3442.0&$3.1 \times 10^6$\\
hepph&9771610&10377853.2&2286.8&$7.2 \times 10^5$&&\textbf{6392653}&\textbf{7055773.8}&3120.6&$2.9 \times 10^6$&&8628687&9370215.3&3376.3&$1.4 \times 10^6$\\
astroph&59029312&60313225.8&3441.4&$5.5 \times 10^5$&&\textbf{55424575}&\textbf{57231348.7}&3576.4&$1.1 \times 10^6$&&62068966&62547898.1&1911.4&$3.9 \times 10^5$\\
condmat&13420836&14823254.9&1481.3&$2.9 \times 10^5$&&\textbf{4086629}&\textbf{5806623.8}&3511.9&$1.6 \times 10^6$&&9454361&10061807.8&1779.5&$4.9 \times 10^5$\\
\midrule[0.5pt]
wins &21.5 & 22.5 & * & * & & 19.0 & 19.5 & * & * & & * & * & * & *\\
\bottomrule[0.75pt]
\end{tabular}
\begin{tablenotes}
\tiny
     \item Note that, it is meaningless to calculate the wins in terms of average time and average steps when different best objective values are achieved, and we represent them by *.
\end{tablenotes}
\end{threeparttable}
\end{scriptsize}
\end{center}
\vskip -0.2in
\end{table*}

From Table \ref{Tab:Comparative Performance of MACNP with Two Best-performing Algorithms CNA1 and FastCNP}, we observe that our MACNP algorithm significantly outperforms CNA1 and FastCNP, achieving the best objective values for 38 out of the 42 instances, and the best average objective values for 37 out of 42 instances. For the synthetic benchmark, MACNP is significantly better than CNA1 in terms of the best objective value, winning 12.5 instances (i.e., $12.5 > CV^{16}_{0.05} = 12$). Compared to FastCNP, MACNP is also very competitive and wins 10.5 instances, which is slightly smaller than the critical value $CV^{16}_{0.05} = 12$. As to the average objective value, MACNP significantly outperforms both CNA1 and FastCNP by winning 13 instances. For the real-world benchmark, MACNP also proves to be significantly better than CNA1 and FastCNP both in terms of the best objective value and the average objective value. Moreover, for the 14 instances where all three algorithms attain the same best objective values, our MACNP algorithm needs the least number of steps to reach its results (see values underlined).

We also compared our MACNP algorithm with five additional algorithms reported in the literature. As the source code of these five reference algorithms is not available, we used their best results reported in the corresponding papers. Fortunately, these five algorithms have been evaluated on the same platform (i.e., an HP ProLiant DL585 G6 server with two 2.1 GHz AMD Opteron 8425HE processors and 16 GB of RAM) \cite{Aringhieri2016b,Aringhieri2016c}, which is slower than our machine with a factor 0.84 according to the Standard Performance Evaluation Corporation (www.spec.org). However, their results were obtained under a longer time limit, i.e., $t_{max} \in (7200,10000]$ for the most of the synthetic instances and $t_{max} \in [3000,16000]$ for most of the real-world instances. Note that, in our comparison, we do not consider the simulated annealing algorithm \cite{Ventresca2012}, the population-based incremental learning algorithm \cite{Ventresca2012}, and the greedy randomized adaptive search procedure with path relinking \cite {Purevsuren2016} because they are completely dominated by FastCNP proposed in \cite{Zhou2017b}.

The comparative results of MACNP with the seven state-of-the-art heuristic algorithms on the synthetic and real-world benchmarks are summarized in Table \ref{Tab:Comparisons of MACNP With the State-of-the-art Algorithms on Synthetic and Realworld Benchmarks}. Note that the result of ``Best ILS'' for each instance is the best result among 6 ILS variants, and ``Best VNS' corresponds to the best result among all 24 VNS variants \cite{Aringhieri2016b}.

\begin{table*}[!ht]
\caption{Comparison between $\rm{MACNP}$ and the state-of-the-art algorithms on synthetic and real-world benchmarks.}
\label{Tab:Comparisons of MACNP With the State-of-the-art Algorithms on Synthetic and Realworld Benchmarks}
\vskip -0.2in
\begin{center}
\begin{scriptsize}
\begin{threeparttable}
\begin{tabular}{lrrrrrrrrrr}
\toprule[0.75pt]
Instance& $K$ & $KBV$ &Greedy3d&Greedy4d&Best VNS&Best ILS&GA&CNA1&FastCNP&MACNP\\
\midrule[0.5pt]
BA500&50&195\tnote{$\ast$}&\textbf{195}&\textbf{195}&\textbf{195}&\textbf{195}&\textbf{195}&\textbf{195}&\textbf{195}&\textbf{195}\\
BA1000 &75&558\tnote{$\ast$}&559&559&559&559&\textbf{558}&\textbf{558}&\textbf{558}&\textbf{558}\\
BA2500 &100&3704\tnote{$\ast$}&3722&3722&\textbf{3704}&3722&\textbf{3704}&\textbf{3704}&\textbf{3704}&\textbf{3704}\\
BA5000 &150&10196\tnote{$\ast$}&\textbf{10196}&\textbf{10196}&\textbf{10196}&10222&\textbf{10196}&\textbf{10196}&\textbf{10196}&\textbf{10196}\\
ER235  &50&295\tnote{$\ast$}&315&313&\textbf{295}&313&\textbf{295}&\textbf{295}&\textbf{295}&\textbf{295}\\
ER466  &80&1524&1938&1993&1542&1874&1560&\textbf{1524}&\textbf{1524}&\textbf{1524}\\
ER941  &140&5012&8106&8419&5198&5544&5120&5114&\textbf{5012}&\textbf{5012}\\
ER2344 &200&959500&1118785&1112685&997839&1038048&1039254&996411&953437&\textbf{902498}\tnote{$\star$}\\
FF250  &50&194\tnote{$\ast$}&199&197&\textbf{194}&195&\textbf{194}&\textbf{194}&\textbf{194}&\textbf{194}\\
FF500  &110&257\tnote{$\ast$}&262&264&\textbf{257}&261&\textbf{257}&263&\textbf{257}&\textbf{257}\\
FF1000 &150&1260\tnote{$\ast$}&1288&1271&\textbf{1260}&1276&\textbf{1260}&1262&\textbf{1260}&\textbf{1260}\\
FF2000 &200&4545\tnote{$\ast$}&4647&4592&4549&4583&4546&4548&4546&\textbf{4545}\tnote{$\star$}\\
WS250  &70&3101&11694&11401&6610&3241&3240&3415&3085&\textbf{3083}\tnote{$\star$}\\
WS500  &125&2078&4818&11981&2130&2282&2199&2085&\textbf{2072}&\textbf{2072}\tnote{$\star$}\\
WS1000 &200&113638&316416&318003&139653&115914&113638&141759&123602&\textbf{109807}\tnote{$\star$}\\
WS1500 &265&13167&157621&243190&13792&14681&13662&13498&13158&\textbf{13098}\tnote{$\star$}\\
\midrule[0.5pt]
wins     &   & 10.5 & 15.0 &15.0  & 12.5 & 15.5 &12.0  &12.5  & 10.5& *\\
\midrule[0.5pt]
Bovine&3&268&\textbf{268}&\textbf{268}&\textbf{268}&\textbf{268}&\textbf{268}&\textbf{268}&\textbf{268}&\textbf{268}\\
Circuit&25&2099&\textbf{2099}&2100&2101&2117&\textbf{2099}&\textbf{2099}&\textbf{2099}&\textbf{2099}\\
E.coli&15&806&\textbf{806}&834&\textbf{806}&\textbf{806}&\textbf{806}&\textbf{806}&\textbf{806}&\textbf{806}\\
USAir97&33&4336&4442&4726&5444&4442&\textbf{4336}&\textbf{4336}&\textbf{4336}&\textbf{4336}\\
HumanDi&52&1115&\textbf{1115}&\textbf{1115}&\textbf{1115}&\textbf{1115}&\textbf{1115}&\textbf{1115}&\textbf{1115}&\textbf{1115}\\
TreniR&26&926&926&936&920&934&928&\textbf{918}&\textbf{918}&\textbf{918}\\
EU\_fli &119&349927&349927&350757&356631&355798&351610&\textbf{348268}&\textbf{348268}&\textbf{348268}\\
openfli&186&28834&29624&29552&31620&29416&28834&29300&28834&\textbf{26842}\tnote{$\star$}\\
yeast1&202&1414&1416&1415&1421&1434&1414&1413&\textbf{1412}&\textbf{1412}\\
H1000&100&328817&338574&336866&332286&344509&328817&314152&314964&\textbf{306349}\tnote{$\star$}\\
H2000&200&1309063&1372109&1367779&1309063&1417341&1315198&1275968&1275204&\textbf{1243859}\tnote{$\star$}\\
H3000a&300&3005183&3087215&3100938&3058656&3278172&3005183&2911369&2885588&\textbf{2844393}\tnote{$\star$}\\
H3000b&300&2993393&3096420&3100748&3121639&3250497&2993393&2907643&2876585&\textbf{2841270}\tnote{$\star$}\\
H3000c&300&2975213&3094459&3097451&3079570&3202002&2975213&2885836&2876026&\textbf{2838429}\tnote{$\star$}\\
H3000d&300&2988605&3090753&3100216&3027839&3237495&2988605&2906121&2894492&\textbf{2831311}\tnote{$\star$}\\
H3000e&300&3001078&3095793&3113514&3031975&3255390&3001078&2903845&2890861&\textbf{2847909}\tnote{$\star$}\\
H4000&400&5403572&5534254&5530402&5498097&5877896&5403572&5194592&5167043&\textbf{5044357}\tnote{$\star$}\\
H5000&500&8411789&8657681&8653358&8889904&9212984&8411789&8142430&8080473&\textbf{7972525}\tnote{$\star$}\\
powergr&494&16099&16373&16406&16099&16533&16254&16158&15982&\textbf{15862}\tnote{$\star$}\\
Oclinks&190&614504&614504&614546&623366&625671&620020&\textbf{611326}&611344&612303\\
faceboo&404&420334&608487&856642&865115&\textbf{420334}&561111&701073&692799&643162\\
grqc&524&13736&13787&13825&13751&13817&13736&15522&13616&\textbf{13596}\tnote{$\star$}\\
hepth&988&114382&232021&326281&114933&123138&114382&130256&108217&\textbf{106397}\tnote{$\star$}\\
hepph&1201&7336826&10305849&10162995&10989642&11759201&7336826&9771610&6392653&\textbf{\textit{6156536}}\tnote{$\star$}\\
astroph&1877&54517114&54713053&54517114&65937108&65822942&58045178&59029312&55424575&\textbf{\textit{53963375}}\tnote{$\star$}\\
condmat&2313&2298596&11771033&11758662&6121430&\textbf{2298596}&2612548&13420836&4086629&\textit{4871607}\\
\midrule[0.5pt]
wins       &   & 22.5 &24.0 & 25.0 & 24.5 & 22.5 & 22.5 & 21.5 & 21.0 & *\\
\bottomrule[0.75pt]
\end{tabular}
\begin{tablenotes}
\tiny
     \item[$\ast$] Optimal results obtained by exact algorithm \cite{Summa2012} within 5 days.
     \item[$\star$] Improved best upper bounds.
     \item Note that, for the results of our MACNP algorithm on hepph, astroph and condmat (in italic format), they are reached with $t_{max} = 16, 000$ seconds.
\end{tablenotes}
\end{threeparttable}
\end{scriptsize}
\end{center}
\vskip -0.2in
\end{table*}

Table \ref{Tab:Comparisons of MACNP With the State-of-the-art Algorithms on Synthetic and Realworld Benchmarks} shows that our MACNP algorithm attains the best results for all instances. Specifically, MACNP finds 6 new upper bounds and reaches the best objective values for the remaining 10 instances. MACNP is significantly better than the reference algorithms except for FastCNP, respectively winning 15.0, 15.0, 12.5, 15.5, 12.0, 12.5 compared to Greedy3d, Greedy4d, Best VNS, Best ILS, GA, and CNA1. Compared to FastCNP,  MACNP wins 10.5 instances, which is just slightly smaller than the critical value $CV_{0.05}^{16} = 12$. These observations indicate that compared to the state-of-the-art algorithms, our MACNP algorithm is highly competitive for solving the synthetic instances. 

Similar observations are found on the real-world benchmark in Table \ref{Tab:Comparisons of MACNP With the State-of-the-art Algorithms on Synthetic and Realworld Benchmarks}. MACNP significantly outperforms the reference algorithms, respectively winning 24.0, 24.0, 24.5, 22.5, 23.5, 21.0, 21.0 instances with respect to the reference algorithms Greedy3d, Greedy4d, Best VNS, Best ILS, GA, CNA1, and FastCNP. Specifically, MACNP achieves the best objective values for 21 out of the 26 real-world instances, including 13 new upper bounds and 8 known best objective values.

\section{Application to the cardinality-constrained critical node problem}
\label{Sec:Application to the Cardinality-Constrained Critical Node Problem}

In this section, we show that our algorithm can also be used to solve other critical node problem, by testing  MACNP on the cardinality-constrained critical node problem (CC-CNP). The experiments were again conducted on the synthetic and real-world benchmarks described in Section \ref{SubSec:Benchmark Instances}.

\subsection{Cardinality-constrained critical node problem}
\label{SubSec:Cardinality-constrained Critical Node Problem}

CC-CNP is a cardinality constrained version of the classic CNP \cite{Arulselvan2011}. CC-CNP aims to identify a minimum subset $S \subseteq V$ such that any connected component in the residual graph $G[V\setminus S]$ contains at most $W$ nodes where $W$ is a given threshold value. 

To approximate $K^*$, we solve a series of CC-CNP with decreasing $K>K^*$ values. For a fixed $K$, we try to find a set $S \subseteq V$ of $K$ nodes, whose removal minimizes the number of nodes in each connected component which exceeds the cardinality threshold $W$. For this purpose, we define an auxiliary (minimization) function $f'$:

\begin{equation} \label{Equ:CC-CNP Objective Function}
	f'(S) = \sum_{i=1}^T \max(|\mathcal{C}_i|-W,0)
\end{equation}

which calculates the total number of nodes in excess of $W$ in all $T$ connected components of the residual graph. It is clear that if $f'(S) = 0$, then $S$ is a feasible solution of CC-CNP.

\subsection{Solving CC-CNP with MACNP}
\label{SubSec:Memetic Algorithm for CC-CNP}

To solve CC-CNP, we adapt MACNP slightly and denote the new algorithm by MACC-CNP. Basically, we replace the objective function $f$ used in MACNP with the minimization function $f'$ defined by Equation (\ref{Equ:CC-CNP Objective Function}). To solve CC-CNP, we start with an initial $K$ value (obtained with a construction method, see below), and apply MACC-CNP to find a set $S$ of $K$ nodes whose removal minimizes the number of exceeded nodes (i.e., minimizing $f'(S)$). If $f'(S) = 0$, then $S$ is a feasible solution of CC-CNP. At this moment, we decrease $K$ by one and solve the problem again. We repeat this process until no feasible solution can be found and report the last $S$ found with  $f'(S) = 0$. This general solution procedure is inspired by a popular approach for solving the classic graph coloring problem \cite{Lu2010, Zhou2016}. 

The initial $K$ value is obtained with an initial feasible solution $S^0$. We first set $S^0$ to be empty. Then we iteratively pick a  node $v$ from a large connected component whose cardinality exceeds $W$ and move $v$ to $S^0$. We repeat this process until a feasible solution $S^0$ is obtained (i.e., $f'(S^0) = 0$, meaning that all components contain at most $W$ nodes). We set the initial $K$ to equal $|S^0|$. 

\subsection{Comparison with the state-of-the-art algorithms}
\label{SubSec:Comparison With The State-of-the-art Algorithms}

Based on the synthetic and real-world benchmarks, we compared our MACC-CNP algorithm with the five state-of-the-art algorithms:  greedy algorithms (G1 and G2) \cite{Aringhieri2016c}, genetic algorithm (GA) \cite{Aringhieri2016c}, multi-start greedy algorithm (CNA2) \cite{Pullan2015}, fast heuristic (FastCNP) \cite{Zhou2017b}. Among these reference algorithms, FastCNP was originally proposed for  the classic critical node problem, and we adapted it for solving CC-CNP in the same way as for MACNP. The source code of CNA2 was provided by its author \cite{Pullan2015}. For algorithms G1, G2 and GA, whose source codes are not available, we used the results reported in \cite{Aringhieri2016c}. These results have been obtained with different time limits $t_{max} \in[100,16000]$ seconds, which are, for most instances, larger than our time limit of $t_{max} = 3600$ seconds. For our comparative study, we ran CNA2 and FastCNP with their default parameters under the time limit $t_{max} = 3600$ seconds, and each instance was solved 30 times. For our MACC-CNP algorithm, we also solved each instance 30 times independently under the same time limit.

The comparative results of running our MACC-CNP algorithm against the reference algorithms on the synthetic and real-world benchmarks are displayed in Table \ref{Tab:Comparison of MACC-CNP With the State-of-the-art Algorithms on Synthetic and Realworld Benchmarks}. To analyze these results, we calculated the number of instances (wins) in which MACC-CNP proved superior according to the two-tailed sign test \cite{Demvsar2006}, as shown in the last row for each benchmark.

\begin{table}[!htbp]
\caption{Comparison between $\rm{MACC-CNP}$ and the state-of-the-art algorithms on synthetic and real-world benchmarks.}
\label{Tab:Comparison of MACC-CNP With the State-of-the-art Algorithms on Synthetic and Realworld Benchmarks}
\vskip -0.2in
\begin{center}
\begin{scriptsize}
\begin{threeparttable}
\begin{tabular}{lrrrrrrrr}
\toprule[0.75pt]
Instance&$L$&$KBV$&G1&G2&GA&CNA2&FA\tnote{$\circ$}&MA\tnote{$\diamond$}\\
\midrule[0.5pt]
BA500&4&47&\textbf{47}&\textbf{47}&\textbf{47}&\textbf{47}&\textbf{47}&\textbf{47}\\
BA1000&5&61&\textbf{61}&\textbf{61}&\textbf{61}&\textbf{61}&\textbf{61}&\textbf{61}\\
BA2500&10&100&101&\textbf{100}&\textbf{100}&\textbf{100}&\textbf{100}&\textbf{100}\\
BA5000&13&149&154&151&\textbf{149}&\textbf{149}&\textbf{149}&\textbf{149}\\
ER235&7&47&49&50&\textbf{47}&\textbf{47}&\textbf{47}&\textbf{47}\\
ER466&14&81&86&85&81&\textbf{79}&\textbf{79}&\textbf{79}\tnote{$\star$}\\
ER941&25&139&149&152&139&141&\textbf{135}&\textbf{135}\tnote{$\star$}\\\
ER2344&1400&204&252&270&204&194&189&\textbf{185}\tnote{$\star$}\\
FF250&5&48&\textbf{48}&49&\textbf{48}&\textbf{48}&\textbf{48}&\textbf{48}\\
FF500&4&100&102&102&\textbf{100}&\textbf{100}&\textbf{100}&\textbf{100}\\
FF1000&7&142&145&145&\textbf{142}&\textbf{142}&\textbf{142}&\textbf{142}\\
FF2000&12&182&191&187&\textbf{182}&\textbf{182}&\textbf{182}&\textbf{182}\\
WS250&40&73&79&80&72&71&\textbf{70}&\textbf{70}\tnote{$\star$}\\\
WS500&15&126&145&144&126&124&\textbf{123}&\textbf{123}\tnote{$\star$}\\\
WS1000&500&162&195&418&162&180&166&\textbf{157}\tnote{$\star$}\\
WS1500&30&278&339&332&278&273&256&\textbf{254}\tnote{$\star$}\\
\midrule[0.5pt]
wins    &&11.5&14.5&14.5&11.5 &11.0 &9.5 &*\\
\midrule[0.5pt]
Bovine&15&4&\textbf{4}&\textbf{4}&\textbf{4}&\textbf{4}&\textbf{4}&\textbf{4}\\
Circuit&30&24&25&26&\textbf{24}&\textbf{24}&\textbf{24}&\textbf{24}\\
E.coli&20&15&16&\textbf{15}&\textbf{15}&\textbf{15}&\textbf{15}&\textbf{15}\\
USAir97&70&33&34&40&\textbf{33}&\textbf{33}&\textbf{33}&\textbf{33}\\
HumanDi&10&49&51&50&\textbf{49}&\textbf{49}&\textbf{49}&\textbf{49}\\
TreniR&10&28&30&31&28&\textbf{27}&\textbf{27}&\textbf{27}\\
EU\_fli &850&113&127&118&113&113&\textbf{112}&\textbf{112}\\
openfli&140&184&194&206&184&183&\textbf{180}&\textbf{180}\\
yeast1&6&195&202&199&195&\textbf{193}&\textbf{193}&\textbf{193}\\
H1000&800&103&172&151&103&97&95&\textbf{92}\tnote{$\star$}\\
H2000&1600&221&362&313&221&207&195&\textbf{188}\tnote{$\star$}\\
H3000a&2500&279&448&402&279&276&252&\textbf{242}\tnote{$\star$}\\
H3000b&2500&279&456&401&279&270&251&\textbf{244}\tnote{$\star$}\\
H3000c&2500&276&446&404&276&274&250&\textbf{244}\tnote{$\star$}\\
H3000d&2500&276&452&402&276&272&250&\textbf{244}\tnote{$\star$}\\
H3000e&2500&280&455&403&280&270&252&\textbf{244}\tnote{$\star$}\\
H4000&3300&398&651&571&398&388&354&\textbf{347}\tnote{$\star$}\\
H5000&4200&458&745&662&458&459&413&\textbf{410}\tnote{$\star$}\\
powergr&20&428&449&440&428&430&\textbf{397}&\textbf{397}\tnote{$\star$}\\
Oclinks&1100&197&209&200&197&193&193&\textbf{192}\tnote{$\star$}\\
faceboo&450&324&472&821&\textbf{324}&523&375&378\\
grqc&20&480&497&501&480&486&462&\textbf{461}\tnote{$\star$}\\
hepth&70&981&1040&1042&981&1029&955&\textbf{944}\tnote{$\star$}\\
hepph&3600&1228&1416&1572&1228&1103&\textbf{994}&1120\tnote{$\star$}\\
astroph&12000&1322&3284&1769&1322&1364&\textbf{1249}&1329\\
condmat&500&2506&2506&2651&2506&2357&2357&\textbf{2320}\tnote{$\star$}\\
\midrule[0.5pt]
wins     & & 21.5 & 25.5 & 25.0 & 21.5 & 22.5 & 19.0 & * \\
\bottomrule[0.75pt]
\end{tabular}
\begin{tablenotes}
\tiny
     \item[$\star$] Improved best upper bounds.
     \item[$\circ$] We adapted the FastCNP algorithm \cite{Zhou2017b} for solving CC-CNP, and the new algorithm is denoted by FastCC-CNP (FA).
     \item[$\diamond$] The proposed MACC-CNP (MA) algorithm.
\end{tablenotes}
\end{threeparttable}
\end{scriptsize}
\end{center}
\vskip -0.2in
\end{table}

As indicated in Table \ref{Tab:Comparison of MACC-CNP With the State-of-the-art Algorithms on Synthetic and Realworld Benchmarks}, MACC-CNP achieves the best objective values for all synthetic instances, and yielding in particular 2 new upper bounds. At a significance level of 0.05, MACC-CNP is significantly better than G1 and G2. For algorithms GA, CNA2 and FastCNP, MACNP is better  but the differences are not significant, winning 11.5, 11.0 and 9.5 instances. We also observe that MACC-CNP is very effective on the real-world benchmark. For this benchmark, at a significance level 0.05, our MACC-CNP algorithm significantly outperforms all reference algorithms. Specifically, MACC-CNP discovers new upper bounds for 15 instances and reaches the known best upper bounds on 8 out of 11 remaining instances. These observations show that MACC-CNP is also highly competitive compared to state-of-the-art algorithms for solving CC-CNP.

\section{Discussion}
\label{Sec:Discussion}

We now analyze the key ingredients of the proposed MACNP algorithm: the two-phase node exchange strategy and node weighting scheme used in the component-based neighborhood search (Sections \ref{SubSec:Benefit of the two-phase node exchange Strategy} and \ref{SubSec:Effectiveness of the Node Weighting Scheme}) and the backbone-based crossover (Section \ref{SubSec:Rationality Behind Double Backbone-based Crossover}). Based on the classic critical node problem, the experiments were carried out on 4 representative synthetic instances from different families (BA5000, ER941, FF500 and WS250) as well as 4 representative real-world instances (TreniR, H3000a, H4000 and hepth). These instances cover different classes with different sizes and have different levels of difficulties. For each algorithm variant, we ran it on each instance 15 times with a time limit $t_{max} = 3600$ seconds.

\subsection{Benefit of the two-phase node exchange strategy}
\label{SubSec:Benefit of the two-phase node exchange Strategy}

Our component-based neighborhood search  decomposes the exchanging procedure into two phases, i.e., the ``add-phase'' and ``removal-phase'', and performs them separately. To investigate the benefit of the two-phase node exchange strategy, we compare MACNP with an alternative algorithm $\rm{MACNP}_{\rm{0}}$ which uses the conventional two node exchange strategy to swap a node $u \in S$ ($S$ being the current solution) with a node $v \in V \setminus S$. $\rm{MACNP}_{\rm{0}}$ and MACNP share thus the same components except the neighborhood.

\begin{table}[!htbp]
\caption{Comparative performance of MACNP with $\rm{MACNP}_{\rm{0}}$.}
\label{Tab:Comparative Performance on Two-phase Node Exchange Strategy}
\begin{center}
\begin{scriptsize}
\begin{tabular}{lrrr|rrr}
\toprule[0.75pt]
\multicolumn{1}{c}{} & \multicolumn{3}{c}{$\rm{MACNP}_{\rm{0}}$} & \multicolumn{3}{c}{$\rm{MACNP}$}\\
\cmidrule[0.5pt]{2-4} \cmidrule[0.5pt]{5-7}
Instance&$f_{best}$&$f_{avg}$&$\frac{steps}{sec}$&$f_{best}$&$f_{avg}$&$\frac{steps}{sec}$\\
\midrule[0.5pt]
BA5000&\textbf{3083}&3257.7&37.6&\textbf{3083}&\textbf{3101.1}&\underline{59954.5}\\
ER941&\textbf{257}&\textbf{257.0}&134.9&\textbf{257}&\textbf{257.0}&\underline{21445.3}\\
FF500&5014&5171.1&5.8&\textbf{5012}&\textbf{5013.7}&29428.5\\
WS250&10198&10200.8&1.5&\textbf{10196}&\textbf{10196.0}&5708.9\\
TreniR&\textbf{918}&\textbf{918.0}&1322.5&\textbf{918}&\textbf{918.0}&\underline{102275.0}\\
H3000a&3451908&3483709.9&$<0.1$&\textbf{2849170}&\textbf{2883554.5}&3909.4\\
H4000 &6165357&6224768.6&$<0.1$&\textbf{5081209}&\textbf{5144354.2}&2117.6\\
hepth&23964174&24528095.0&$<0.1$&\textbf{106552}&\textbf{108354.0}&1100.8\\
\bottomrule[0.75pt]
\end{tabular}
\end{scriptsize}
\end{center}
\vskip -0.2in
\end{table}

The comparative results for MACNP and $\rm{MACNP}_{\rm{0}}$ are summarized in Table \ref{Tab:Comparative Performance on Two-phase Node Exchange Strategy}. In this table, for each instance, we report the best objective value ($f_{best}$), the average objective value ($f_{avg}$), and the average number of steps per second ($\frac{steps}{sec}$) over 15 trials achieved by each algorithm. The comparative results show that MACNP performs significantly better than $\rm{MACNP}_{\rm{0}}$ in terms of all comparison indicators primarily due to its much lower computational complexity per step. In each second, MACNP is able to perform seventy or even tens of thousands times more steps than that of $\rm{MACNP}_{\rm{0}}$. 

Finally, even if we do not show a direct comparison between our component-based neighborhood with the two neighborhoods proposed in \cite{Aringhieri2016b}, Table \ref{Tab:Comparisons of MACNP With the State-of-the-art Algorithms on Synthetic and Realworld Benchmarks}) (columns 10 and 7) shows that FastCNP (using our component-based neighborhood) clearly dominates the Best ILS algorithm (using the two complementary neighborhoods), which provides an additional indicator of the interest of the proposed component-based neighborhood.


\subsection{Effectiveness of the node weighting scheme}
\label{SubSec:Effectiveness of the Node Weighting Scheme}

To assess the effectiveness of the node weighting scheme in the component-based neighborhood search (CBNS), we compared our MACNP algorithm with its alternative algorithm $\rm{MACNP}_{\rm{1}}$ where the node weighting scheme is disabled in CBNS. In $\rm{MACNP}_{\rm{1}}$, we also decompose the exchanging operation into two phases (``add-phase'' and ``removal-phase''), and execute them separately. However, for ``add-phase'', a node is randomly removed from a large connected component instead of selecting the node by the node weighting scheme.

Table \ref{Tab:Comparative Performance on Weighting Scheme} shows the comparative results of MACNP and $\rm{MACNP}_{\rm{1}}$ on the tested instances, based on four indicators: best objective value ($f_{best}$), average objective value ($f_{avg}$), average time to find the best objective value ($t_{avg}$), average number of steps required to find the objective value ($\#steps$). An obvious observation from this table is that the algorithm with the node weighting scheme (i.e., MACNP) significantly outperforms $\rm{MACNP}_{\rm{1}}$ (which lacks the node weighting scheme) on almost all instances except WS250. For WS250, both MACNP and $\rm{MACNP}_{\rm{1}}$ achieve the best objective value 3083, while the average objective value 3093.8 of $\rm{MACNP}_{\rm{1}}$ is slightly better than 3101.1 of MACNP. More importantly, MACNP needs less time and fewer steps to achieve the best objective values. These observations demonstrate the effectiveness of the node weighting scheme.

\begin{table*}[!hbtp]
\caption{Comparative performance of MACNP with its alternative $\rm{MACNP}_{\rm{1}}$ (without node weighing scheme).}
\label{Tab:Comparative Performance on Weighting Scheme}
\vskip -0.2in
\begin{center}
\begin{scriptsize}
\begin{tabular}{lrrrccrrrc}
\toprule[0.75pt]
\multicolumn{1}{c}{} & \multicolumn{4}{c}{$\rm{MACNP}_{\rm{1}}$} && \multicolumn{4}{c}{$\rm{MACNP}$}\\
\cmidrule[0.5pt]{2-5} \cmidrule[0.5pt]{7-10}
Instance&$f_{best}$&$f_{avg}$&$t_{avg}$&\#steps &&$f_{best}$&$f_{avg}$&$t_{avg}$&\#steps\\
\midrule[0.5pt]
BA5000&\textbf{10196}&\textbf{10196.0}&44.7&$2.7 \times 10^5$&&\textbf{10196}&\textbf{10196.0}&5.8&\underline{$3.3 \times 10^4$}\\
ER941&5014&5014.4&310.1&$9.3 \times 10^6$&&\textbf{5012}&\textbf{5013.7}&612.5&$1.8 \times 10^7$\\
FF500&\textbf{257}&\textbf{257.0}&0.5&$1.2 \times 10^4$&&\textbf{257}&\textbf{257.0}&0.3&\underline{$6.4 \times 10^3$}\\
WS250&\textbf{3083}&\textbf{3093.8}&1360.2&$8.5 \times 10^7$&&\textbf{3083}&3101.1&1185.6&\underline{$7.1 \times 10^7$}\\
TreniR&\textbf{918}&\textbf{918.0}&0.3&$3.5 \times 10^4$&&\textbf{918}&\textbf{918.0}&0.2&\underline{$2.0 \times 10^4$}\\
H3000a&2862217&2913716.6&3202.4&$1.2 \times 10^7$&&\textbf{2849170}&\textbf{2883554.5}&3270.2&$1.3 \times 10^7$\\
H4000 & 5187236 & 5313477.9 & 3202.0 & $9.0 \times 10^6$ &&\textbf{5081209} & \textbf{5144354.2} & 2985.9 & $6.3 \times 10^6$\\
hepth&108453&111804.3&3431.6&$3.3 \times 10^6$&&\textbf{106552}&\textbf{108354.0}&3369.1&$3.7 \times 10^6$\\
\bottomrule[0.75pt]
\end{tabular}
\end{scriptsize}
\end{center}
\vskip -0.1in
\end{table*}

\subsection{Rationale behind the double backbone-based crossover}
\label{SubSec:Rationality Behind Double Backbone-based Crossover}

As introduced in Section \ref{SubSec:Double Backbone Based Crossover}, we specially designed a double backbone-based crossover operator to generate offspring solutions. To investigate the rationale behind this crossover, we compare MACNP with an alternative version $\rm{MACNP}_{\rm{2}}$. $\rm{MACNP}_{\rm{2}}$ is obtained from MACNP by replacing our dedicated backbone-based crossover with a single backbone-based crossover which only treats the common elements as the backbone. Specifically, the single backbone-based crossover operator first constructs a partial solution $S^0$ by directly inheriting all the common elements of two parent solutions and then completes the partial solution $S^0$ by removing a node from a large component of the residual graph until $|S^0| = K$.

\begin{figure}[!htbp]
\centering
\includegraphics[width=3.3in]{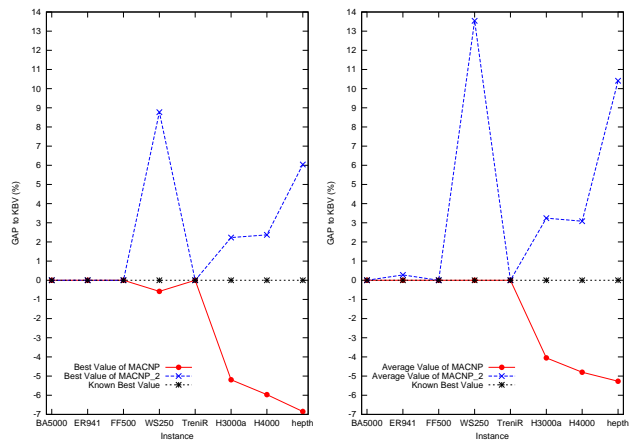}
\caption{Comparative results of MACNP and its alternative $\rm{MACNP}_{\rm{2}}$.}
\label{Fig:Comparative Results of Different Crossover Operators}
\end{figure}

Comparative performance of MACNP and its alternative $\rm{MACNP}_{\rm{2}}$ in terms of the best objective value and average objective value are displayed in the left and right part of Figure \ref{Fig:Comparative Results of Different Crossover Operators} respectively. The X-axis indicates the instance, and the Y-axis shows the gap between our results (eight best values or average values) to the known best values in percentage, which is defined as $(f-KBV)\times 100/KBV$ where $f$ is the best objective value or average objective value, and $KBV$ is the known best objective (see 3rd column of Table \ref{Tab:Comparisons of MACNP With the State-of-the-art Algorithms on Synthetic and Realworld Benchmarks}). A gap smaller than zero means the algorithm obtains a new upper bound for the corresponding instance.

From Figure \ref{Fig:Comparative Results of Different Crossover Operators}, we observe that compared to $\rm{MACNP}_{\rm{2}}$, our MACNP algorithm is able to attain a better $f_{best}$ for all 8 instances including 4 new upper bounds (see values below 0 on the left part  of Figure \ref{Fig:Comparative Results of Different Crossover Operators}). MACNP also outperforms $\rm{MACNP}_{\rm{2}}$ on all 8 tested instances in terms of the average objective value ($f_{avg}$), as shown in the right part of Figure \ref{Fig:Comparative Results of Different Crossover Operators}. This experiment confirms the value  of our double backbone-based crossover.

\section{Conclusions and future work}
\label{Sec:Conclusions and Future Work}

In this work, we proposed an effective memetic search approach for solving the classic critical node problem (MACNP), which combines a component-based neighborhood search for local optimization, a double backbone-based crossover operator for solution recombination and a rank-based pool updating strategy to guarantee a healthy diversity of the population. To ensure its effectiveness, the component-based neighborhood search relies on a focused and reduced neighborhood owing to its two-phase node exchange strategy and the node weighting scheme. Additionally, the double backbone-based crossover not only conserves solution features from the parent solutions, but also introduces diversity by including exclusive elements from parent solutions in a probabilistic way.

To demonstrate the competitiveness of the proposed algorithm, we evaluated MACNP on a broad range of synthetic and real-world benchmarks. The computational results showed that MACNP significantly outperforms state-of-the-art algorithms on both two benchmarks. We also assessed the performance of MACNP for solving the cardinality-constrained critical node problem (MACC-CNP), which is an important variant of the classic CNP. Our results showed that the approach is also highly competitive compared with state-of-the-art algorithms. Finally, we performed experiments to investigate the benefit of different search components and techniques.

Future work motivated by our findings will be to investigate opportunities for further improving the performance of our algorithm by incorporating machine learning techniques (e.g. reinforcement learning and opposition-based learning). Another inviting avenue for research is to adapt the proposed approach to solve other critical node problems with different measures (e.g. distance-based connectivity and betweenness centrality).

\section*{Acknowledgment}
We would like to thank Dr. W.~Pullan for kindly sharing the source codes of the CNA1 and CNA2 algorithms. 

\ifCLASSOPTIONcaptionsoff
  \newpage
\fi



\bibliographystyle{IEEEtran}
\end{document}